\documentclass{article}

\usepackage{microtype}
\usepackage{graphicx}
\usepackage{subcaption}
\usepackage{booktabs} 
\usepackage{dsfont}
\usepackage{enumitem}
\usepackage{hyperref}

\usepackage[preprint]{icml2026}

\usepackage{amsmath}
\usepackage{amssymb}
\usepackage{mathtools}
\usepackage{amsthm}
\usepackage{booktabs} 
\usepackage{caption}
\usepackage{makecell}
\usepackage{multirow}
\usepackage[capitalize,noabbrev]{cleveref}

\theoremstyle{plain}

\theoremstyle{definition}

\theoremstyle{remark}

\usepackage[textsize=tiny]{todonotes}

\icmltitlerunning{DTS: Enhancing Large Reasoning Models via Decoding Tree Sketching}

\begin{document}
\definecolor{darkred}{rgb}{0.55, 0.0, 0.0}
\definecolor{lightgreen}{HTML}{41bf52}
\definecolor{lightpurple}{HTML}{d184c9}
\def\Algname{Decoding Tree Sketching}
\def\Algnameunderline{\textbf{D}ecoding \textbf{T}ree \textbf{S}ketching}
\def\Algnametitleabbr{DTS}
\def\Algnameabbr{\texttt{DTS}}
\definecolor{lightgreen}{RGB}{144,238,144} 
\definecolor{lightblue}{HTML}{4884bd}  
\definecolor{lightred}{RGB}{255,160,160}   

\twocolumn[
  \icmltitle{DTS: Enhancing Large Reasoning Models via \\
             Decoding Tree Sketching}



  \icmlsetsymbol{equal}{*}

  \begin{icmlauthorlist}
    \icmlauthor{Zicheng Xu}{equal,jhu}
    \icmlauthor{Xiuyi Lou}{equal,jhu}
    \icmlauthor{Guanchu Wang$\text{ }{}^ \dagger$}{equal,uncc}
    \icmlauthor{Yu-Neng Chuang}{rice}
    \icmlauthor{Feng Luo}{rice}
    \icmlauthor{Guangyao Zheng}{jhu}
    \icmlauthor{Alexander S. Szalay}{jhu}
    \icmlauthor{Zirui Liu}{umn}
    \icmlauthor{Vladimir Braverman$\text{ }{}^ \dagger$}{jhu}
  \end{icmlauthorlist}

  \icmlaffiliation{jhu}{Department of Computer Science, Johns Hopkins University, Baltimore, USA}
  \icmlaffiliation{uncc}{Department of Computer Science,  University of North Carolina at Charlotte, Charlotte, USA}
  \icmlaffiliation{rice}{Department of Computer Science, Rice University, Houston, USA}
  \icmlaffiliation{umn}{Department of Computer Science, University of Minnesota, 
  Minneapolis, USA}

  \icmlcorrespondingauthor{Vladimir Braverman}{vova@cs.jhu.edu}
  \icmlcorrespondingauthor{Guanchu Wang}{gwang16@charlotte.edu}

  \icmlkeywords{Machine Learning, ICML}

  \vskip 0.3in
]



\printAffiliationsAndNotice{}  

\begin{abstract}
Large Reasoning Models (LRMs) achieve remarkable inference-time improvements through parallel thinking. 
However, existing methods rely on redundant sampling of reasoning trajectories, failing to effectively explore the reasoning space to uncover high-quality solutions.
To address these limitations, we propose \Algname{} (\Algnameabbr{}), a plug-and-play decoding framework for structural multi-trajectory exploration and reasoning selection. 
For reasoning exploration, \Algnameabbr{} sketches a backbone tree of the reasoning space by selectively branching at decision tokens. For reasoning selection, guided by length-accuracy anti-correlation, \Algnameabbr{} designs an early termination to prioritize short and reliable trajectories during decoding.
Experimental results across four LRMs and datasets demonstrate that \Algnameabbr{} significantly enhances accuracy by \textbf{14\%} and reduces repetitive generation by \textbf{8\%} on average. Notably, \Algnameabbr{} enables smaller models to outperform larger models with 10$\times$ the size, highlighting its potential to strengthen reasoning capabilities. The source code is available at: \url{https://github.com/ZichengXu/Decoding-Tree-Sketching}.

\end{abstract}
\vspace{-5mm}
\section{Introduction}

Large Reasoning Models (LRMs), such as DeepSeek-R1~\cite{guo2025deepseek} and Qwen3~\cite{yang2025qwen3}, have demonstrated impressive reasoning capabilities across domains such as mathematics, programming, and scientific problem-solving~\cite{zhang2025survey, chuang2025confident, xu2025self}. These models have shown strong inference-time reasoning performance on challenging tasks~\cite{leang2024comat, zhou2024self} by generating explicit, step-by-step chains of thought (CoT)~\cite{wei2022chain}. Beyond single reasoning trajectory, recent work has shown that reasoning through parallel thinking can substantially enhance LRM performance. Methods such as Self-Consistency~\cite{wang2022self} and DeepConf~\cite{fu2025deep} generate multiple reasoning trajectories and aggregate their results, allowing models to offset local errors and consider alternative solutions. Empirically, such multi-trajectory decoding often outperforms single-trajectory inference, highlighting parallel thinking as a powerful mechanism for improving reasoning accuracy.

\begin{figure}
    \centering
    \captionsetup{belowskip=-2mm}
    \!\!\!\!\!\!\!\!\includegraphics[width=\linewidth]{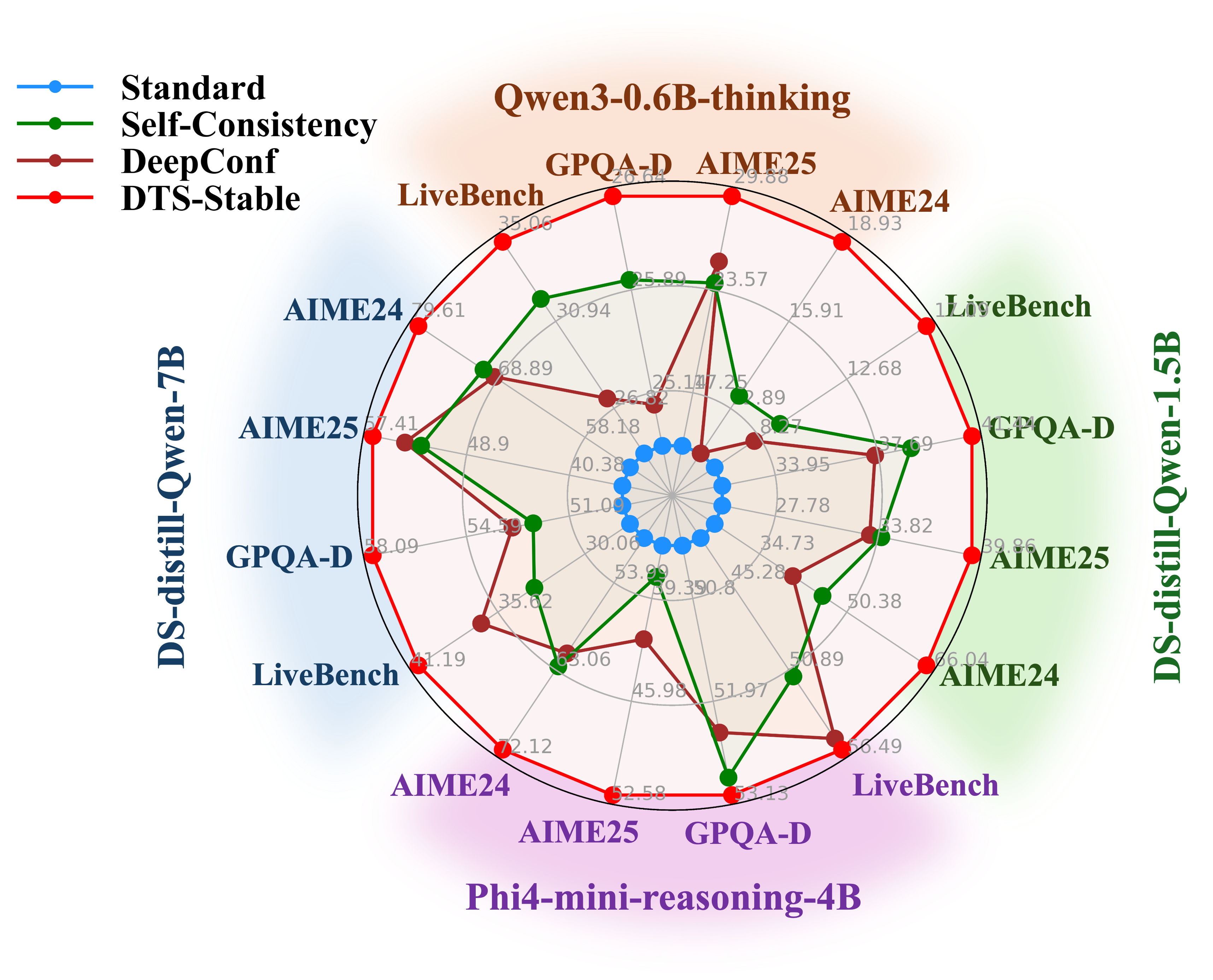}
    \caption{Comprehensive performance of \Algnameabbr{} on the AIME24, AIME25, GPQA-D, LiveBench datasets.}
    \label{fig:radar}
\end{figure}

However, existing parallel thinking approaches exhibit two fundamental limitations. First, most methods independently generate reasoning trajectories. The parallel reasoning processes are often highly identical and redundant, failing to provide meaningful semantic exploration of the underlying reasoning space. As a result, increasing the number of samples does not necessarily lead to better coverage or diversity of reasoning trajectories. Second, trajectory selection is typically applied only post hoc, after all reasoning sequences have fully completed, providing insufficient robust principles to guide or prioritize promising reasoning trajectories during decoding. Together, these limitations suggest that effective parallel thinking requires both \textbf{structured exploration} and \textbf{reliable trajectory selection}. 
To address these limitations, we propose \Algnameunderline{} (\Algnameabbr{}) that sketches a structured reasoning tree and prioritizes high-quality solutions via length-aware early termination within a unified decoding framework.

\subsection*{How to structurally explore the reasoning space?}
Exploring multiple reasoning trajectories can be viewed as searching over an implicit reasoning space. LRM stochastic decoding induces a space of alternative reasoning trajectories that can be naturally organized as a tree structure, where each node corresponds to a generated token and each root-to-leaf path represents a complete CoT reasoning trajectory. Conceptually, an oracle that could exhaustively enumerate and evaluate all paths in this tree would identify the best reasoning trajectory and recover the correct answer. However, in practice, the exponential growth of this tree produces a practically infinite search space, rendering exhaustive enumeration computationally infeasible. 
Therefore, our proposed \Algnameabbr{} \textbf{sketches} the reasoning space into a compact backbone tree that preserves critical branching structure while avoiding full enumeration. \Algnameabbr{} adopts next-token varentropy and entropy to identify decision tokens where several semantically plausible continuations exist. By selectively expanding branches at these tokens, a dynamically sketched reasoning tree is constructed, enabling structural exploration of the reasoning space.
\vspace{-1mm}
\subsection*{Is there any insight to support the selection?}
\vspace{-1mm}

Our analysis in Section~\ref{sec:overthinking} reveals an \textbf{anti-correlation} between reasoning length and task accuracy. Empirically, we observe that high-quality solutions often lie along relatively short trajectories within the reasoning tree. We also provide a theoretical foundation for this anti-correlation by examining the Group Relative Policy Optimization (GRPO)~\cite{shao2024deepseekmath} training objective, which introduces an asymmetric bias that favors short successful reasoning and long unsuccessful ones. Building upon this insight, \Algnameabbr{} proposes an early termination criteria that prioritizes the earliest completed trajectories for the final solution. This allows promising solutions to be identified directly during decoding, aligning trajectory selection with the anti-correlation to favor short, reliable trajectories.
\vspace{-1mm}
\subsection*{How does \Algnameabbr{} perform in practice?}
\vspace{-1mm}
We evaluate \Algnameabbr{} on four LRMs across four reasoning datasets. As demonstrated in Figure~\ref{fig:radar}, \Algnameabbr{} attains accuracy improvements of \textbf{14\%} on average. Our contributions are summarized as follows:
\vspace{-1mm}
\begin{itemize}[leftmargin=10pt]
    \itemsep=-1pt
    \item \textbf{Training-free decoding:} \Algnameabbr{} requires no training, operates entirely at decoding time, and serves as a plug-and-play module for general LRMs.
    \item \textbf{Structured tree sketching:} \Algnameabbr{} constructs a dynamic reasoning tree by selectively branching at decision tokens, enabling the exploration of multiple diverse reasoning trajectories within a structured search space.
    \item \textbf{Reliable selection:} \Algnameabbr{} leverages the anti-correlation between accuracy and reasoning length to select short yet reliable trajectories for final solution.
    \item \textbf{Evaluation:} Across four LRMs and datasets, \Algnameabbr{} consistently improves accuracy and reduces repetition.
\end{itemize}

\vspace{-3mm}

\begin{figure*}[t]
    \vspace{-3mm}
    \centering
    \begin{subfigure}[t]{0.33\textwidth}
        \centering
        \includegraphics[width=\linewidth]{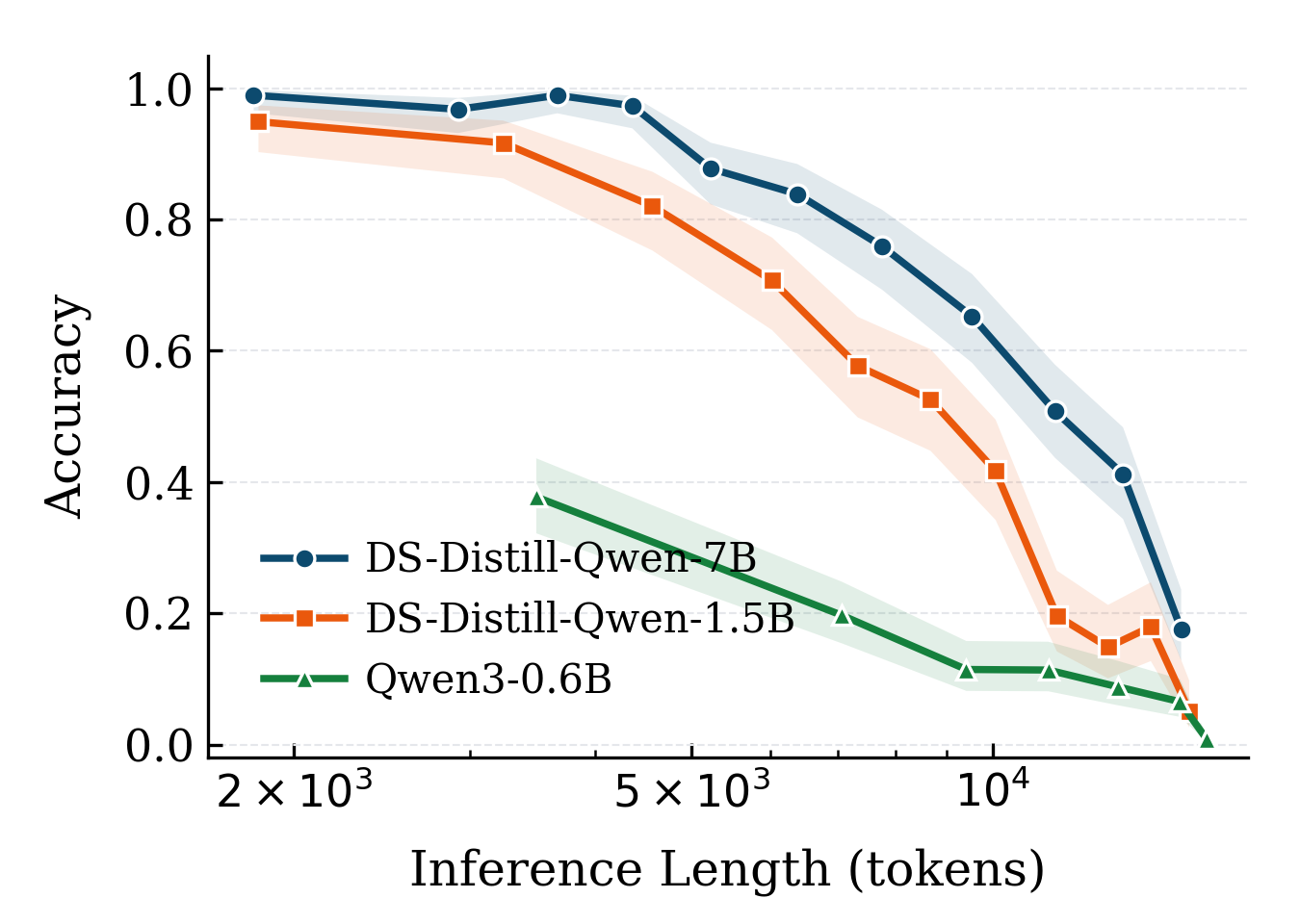}
        \caption{AIME24} 
    \end{subfigure}
    \begin{subfigure}[t]{0.33\textwidth}
        \centering
        \includegraphics[width=\linewidth]{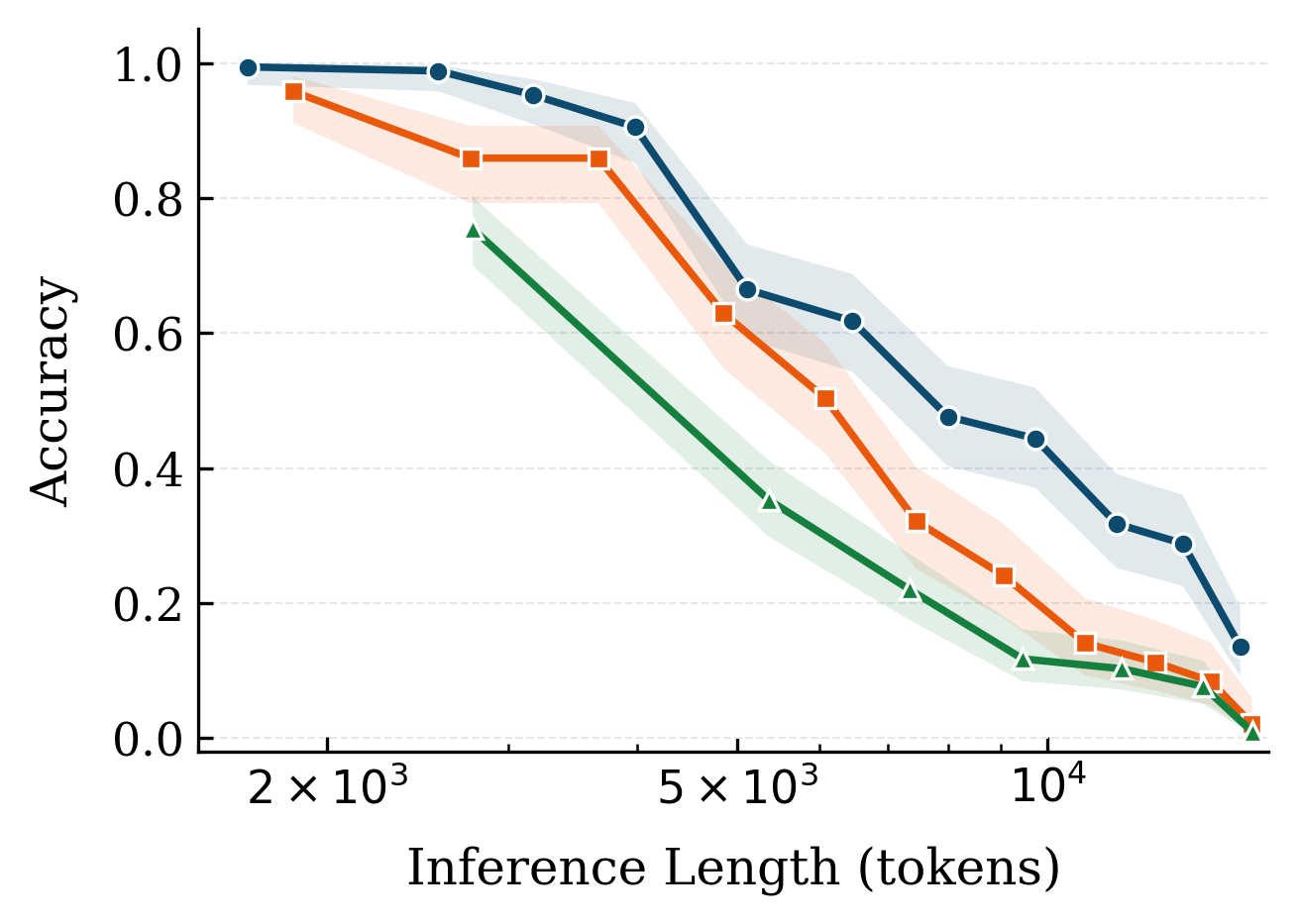}
        \caption{AIME25}
    \end{subfigure}
    \begin{subfigure}[t]{0.33\textwidth}
        \centering
        \includegraphics[width=\linewidth]{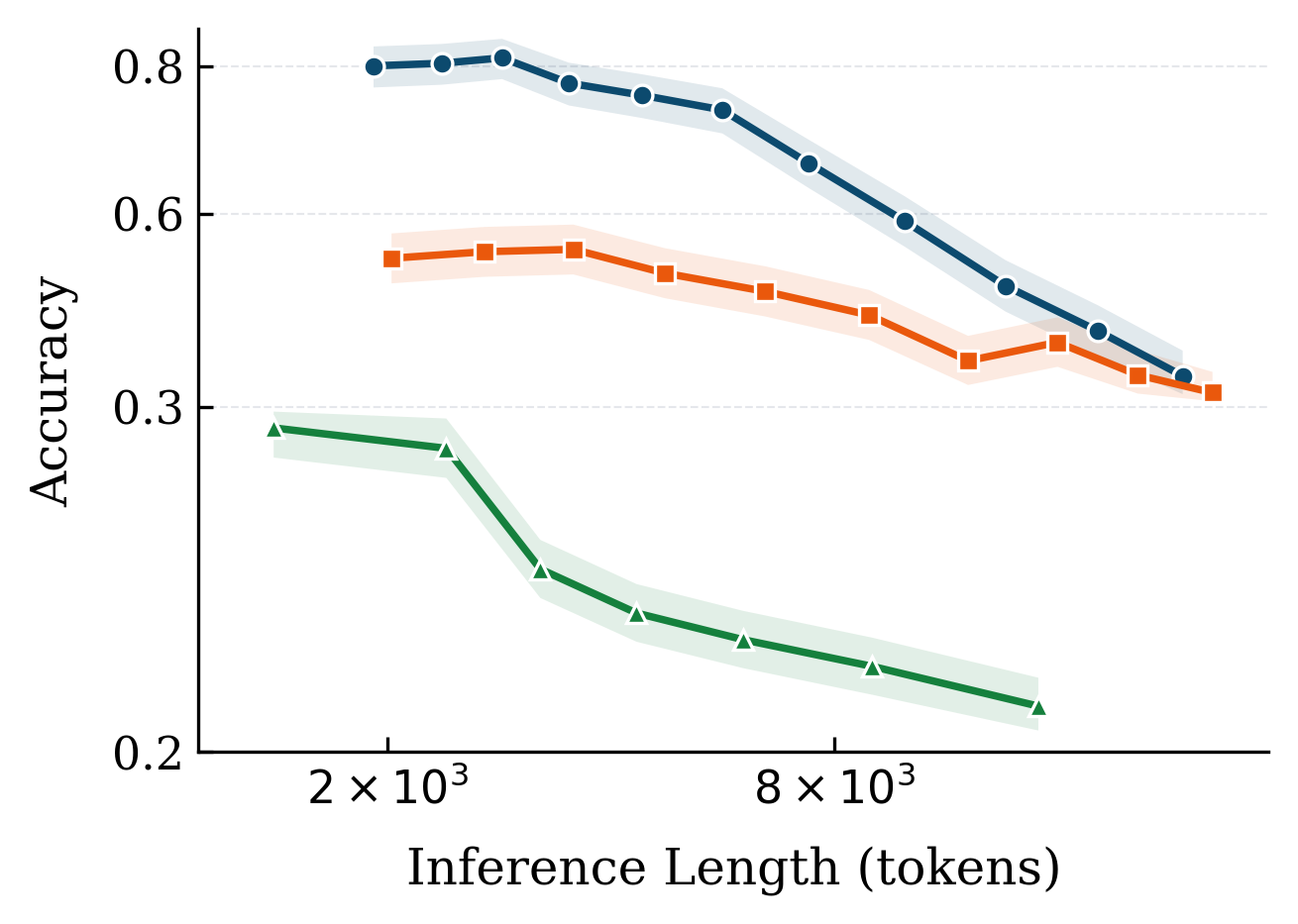}
        \caption{GPQA-D} 
    \end{subfigure}
    \caption{Demonstration of the anti-correlation between accuracy and reasoning length. Experiments are conducted with the DS-Distill-Qwen-1.5B, 7B and Qwen3-0.6B LRMs on the (a) AIME24, (b) AIME25, and (c) GPQA-D datsets.}
    \label{fig:anti-correlation}
    \vspace{-3mm}
\end{figure*}
\vspace{-1mm}
\section{Preliminary}
\subsection{Notations}
We consider LRM $f$ in this work.
For each input prompt $x$, the reasoning trajectory is auto-regressively generated by $\xi_{t+1} = \xi_t \oplus v_t$, and $v_t \sim f(x_t, \xi_t)$ for $t = 1, 2, \cdots$, where $\xi_t \oplus v_j$ denotes to attach token $v_j$ to trajectory $\xi_t$.
\vspace{-1mm}
\subsection{Length-Accuracy Anti-correlation}
\label{sec:overthinking}
In this section, we reveal a \textbf{length-accuracy anti-correlation} for LRMs, where shorter reasoning trajectories consistently achieve higher task accuracy than longer ones.
\vspace{-1mm}
\paragraph{Empirical Evidence.} 
We comprehensively validate the length-accuracy anti-correlation on the AIME24~\cite{aime24}, AIME25~\cite{aime25}, and GPQA-D~\cite{rein2024gpqa} datasets using DeepSeek-Distill-Qwen-1.5B, -7B~\cite{guo2025deepseek}, and Qwen3-0.6B~\cite{yang2025qwen3}. For each problem, we sample 100 stochastic reasoning trajectories.
We partitioned these trajectories into groups based on their length and calculated the mean accuracy for each group. Figure~\ref{fig:anti-correlation} illustrates this relationship, where each point represents the average accuracy for a specific group. Across all cases, we observe a clear anti-correlation with accuracy decreases as length increases, indicating that shorter reasoning chains are more reliable. This length-accuracy anti-correlation motivates \Algnameabbr{}'s objective to select the shortest reasoning trajectories in the reasoning space and enhance LRMs' performance. Beyond its empirical validation, this anti-correlation is consistent with the Reinforcement Learning (RL) objectives used to train LRMs.
\vspace{-1mm}
\paragraph{Theoretical Foundation.} 
Recent studies have shown that RL post-training introduces a systematic length bias in LRMs~\cite{liu2025understanding, devic2025trace}. 
Under GRPO objectives, each generated reasoning trajectory is optimized by aggregating token-level updates weighted by a normalized advantage. 
The abbreviated GRPO objective\footnote{The full GRPO objective can be referred to Equation (1) in the DeepSeek-R1 technical report~\cite{guo2025deepseek}.} can be written as 
\begin{equation}
\label{eq:grpo_abbrev_simple}
\mathcal{J}_{GRPO}(f)
=
\mathbb{E}_{\mathcal{G}\sim f}
\left[
\frac{1}{|\mathcal{G}|}\sum_{\xi\in\mathcal{G}}
\frac{1}{|\xi|}\sum_{t=1}^{|\xi|}
\ell_t \,\hat A_{t}
\right],
\end{equation}
where $\mathcal{G}$ is the group of sampled trajectories, $\ell_t$ is the clipped PPO term, and $\hat A_{t}$ is the normalized advantage assigned to the trajectory $\xi$.
The length normalization by $|\xi|$ in the equation induces an asymmetric learning bias.
For correct reasoning with a positive advantage ($\hat A_{t}>0$), shorter trajectories receive larger effective updates.
Conversely, for incorrect reasoning with a negative advantage ($\hat A_{t}<0$), longer trajectories reduce the penalty.
Therefore, this bias implicitly motivates LRMs to correlate correctness with brevity and incorrectness with verbosity, providing a theoretical foundation for an anti-correlation between reasoning length and accuracy in LRMs.

\subsection{Chasing Shortest Reasoning Trajectories by Decoding Tree}
We follow the anti-correlation between the accuracy and reasoning length to optimize the reasoning process. 
To represent the reasoning space, all possible reasoning trajectories of an LRM can be naturally represented as a tree structure, where each node corresponds to a possible token in the generated trajectory. 
Starting from the first token, every step in the reasoning process branches into $|\mathbb{T}|$ possible continuations, where $\mathbb{T}$ denotes the token space. 
In the following steps, the second token branches into $|\mathbb{T}|^2$ possible paths, the third token into $|\mathbb{T}|^3$, and so forth, leading to an exponentially growing tree space $|\mathbb{T}| + |\mathbb{T}|^2 + |\mathbb{T}|^3 + \cdots$.

According to the anti-correlation, shorter reasoning trajectories generally achieve higher accuracy, suggesting that the \textbf{optimal solution lies in identifying the shortest paths from the root token to a leaf node}. 
However, the exponential explosion of the reasoning tree produces an effectively infinite search space, making it computationally infeasible to exhaustively traverse every path for finding the globally optimal reasoning trajectory.
To this end, our \Algnameabbr{} can effectively prune the search space by sketching the growing tree space during decoding, and approximates the global optimal solution by selecting the shortest reasoning trajectories, rather than exhaustive full enumeration.

\begin{figure*}[t]
    \vspace{-1mm}
    \centering
    \begin{subfigure}[t]{0.5\textwidth}
        \centering
        \includegraphics[width=\linewidth]{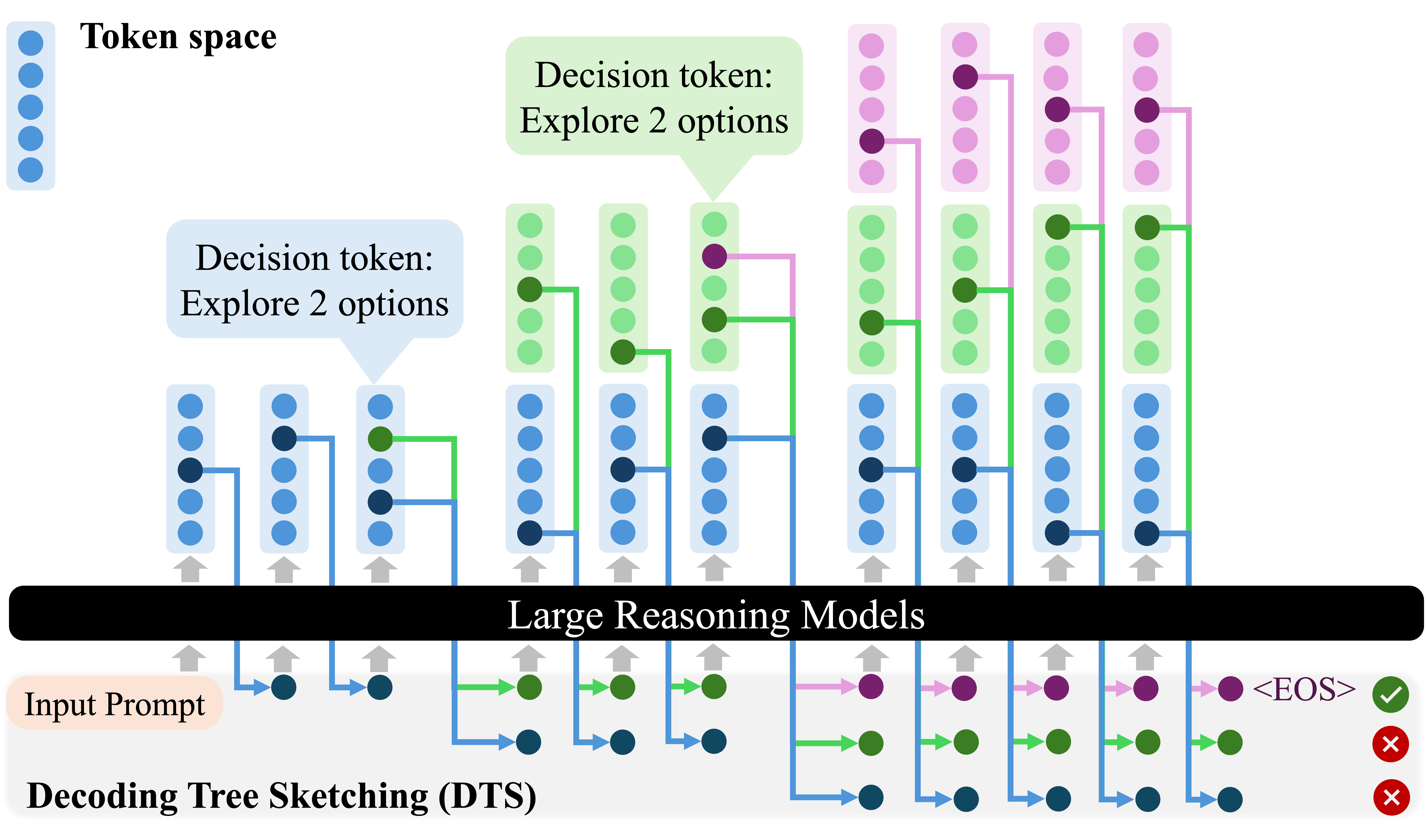}
        \caption{}
    \end{subfigure}
    \quad
    \begin{subfigure}[t]{0.45\textwidth}
        \centering
        \includegraphics[width=\textwidth]{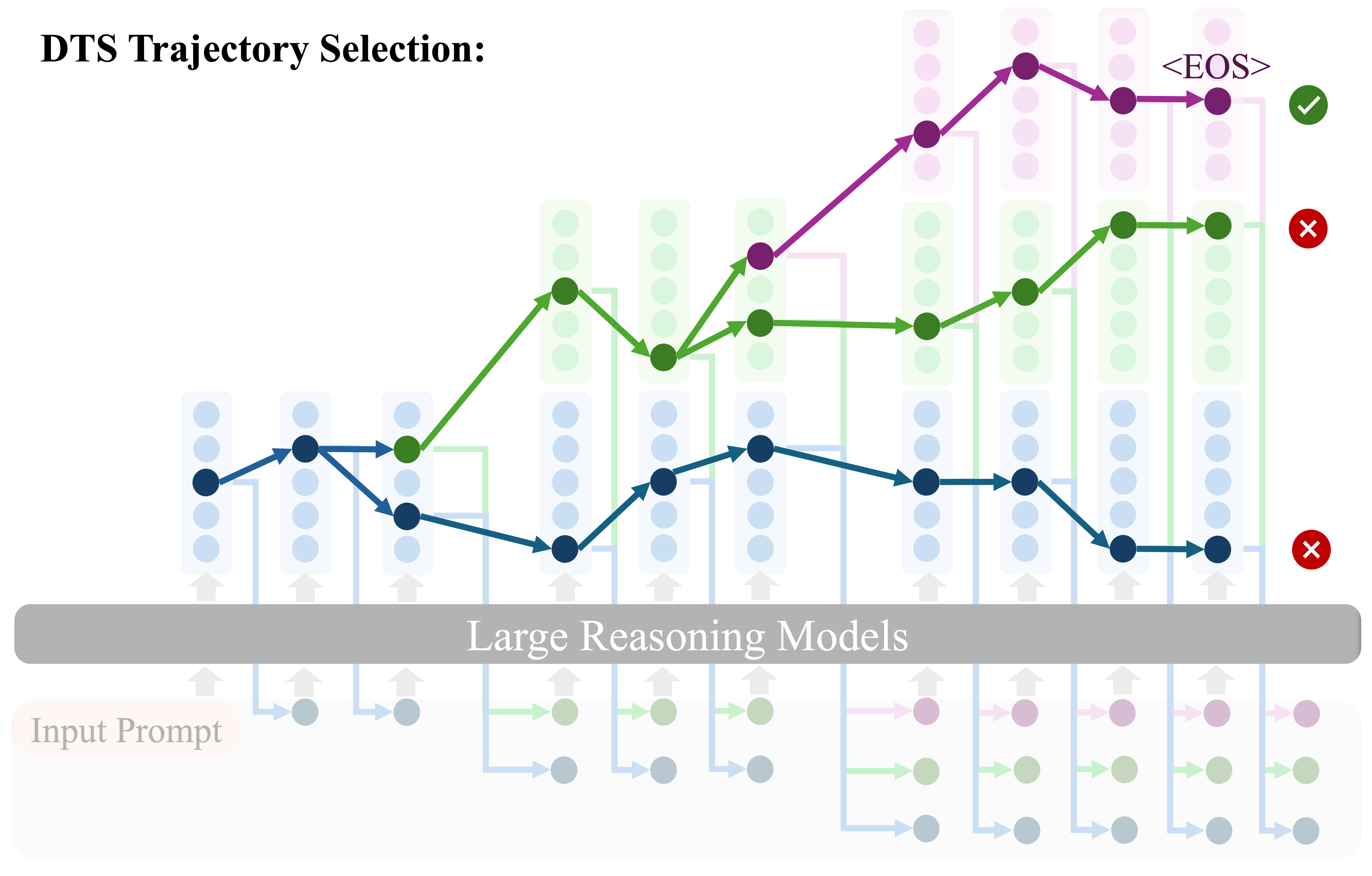}
        \caption{}
    \end{subfigure}
    \caption{\label{fig:acc-vs-tokens}(a) Generation of the decoding tree by \Algnameabbr{}. Starting from the input prompt, \Algnameabbr{} first follows standard auto-regressive decoding, producing a single branch (\textcolor{lightblue}{blue}). At decision tokens, \Algnameabbr{} generates new branches (\textcolor{lightgreen}{green} and \textcolor{lightpurple}{purple}) by selecting top-$K$ candidate tokens. Each branch expands in parallel across its own token space. (b) \Algnameabbr{} trajectory selection via early termination. All branches share prefix computation up to their branching points. \Algnameabbr{} prioritizes the earliest completed branches that reach the end-of-sequence token (\textcolor{lightpurple}{purple}). }
    \label{fig:dts-framework}
    \vspace{-3mm}
\end{figure*}

\section{\Algname~(\Algnameabbr{})}
We introduce \Algnameabbr{} to sketch the decoding tree and approximate the optimal solution.
The overall framework of \Algnameabbr{} is illustrated in Figure~\ref{fig:dts-framework}.
\Algnameabbr{} consists of two key modules: decoding tree generation and early termination strategy.
Together, these modules enable structured exploration of the reasoning tree and reliable trajectory selection, allowing \Algnameabbr{} to select the optimal reasoning path.

\subsection{Decoding Tree Generation}
\Algnameabbr{} constructs the decoding tree with two key components: selective branch generation and parallel branch expansion.
These components define how the tree grows and decodes trajectories simultaneously.
We describe each component in detail below.
\vspace{-1mm}
\paragraph{Selective Branch Generation.}
Unlike standard auto-regressive decoding, which generates a single token at each step, \Algnameabbr{} adaptively expands branches when a decision token is encountered, where several semantically distinct continuations remain plausible. Specifically, a decision token's next-token distribution should have several preferred tokens, each carrying noticeable probability mass and corresponding to different reasoning directions. To identify these decision tokens, \Algnameabbr{} examines entropy and varentropy of the next-token distribution. 
Entropy reflects the concentration of probability mass and decreases when the distribution is highly peaked at a few high-probability tokens. Varentropy measures the variance of the information content and increases when the distribution exhibits a high disparity in the surprisals of its top candidates, signaling a competition between different reasoning directions. Therefore, decision tokens are characterized by low entropy and high varentropy where the model commits strongly to a small yet semantically distinct set of continuations, making branching especially valuable.
Formally, given an input prompt $x$ and an intermediate reasoning trajectory $\xi_t$, let
$P_t(v) = f(x,\xi_{t})$ denote the next-token distribution produced by the LRM at step $t$. 
We first define a decision token criteria $\delta_{\text{dec}}(P_t)$ that determines whether a token is a decision token:
\begin{equation}
\delta_{\text{dec}}(P_t) =
\begin{cases}
1 & \text{if } \mathrm{VE}(P_t)\ge\tau_{\mathrm{v}} \ \text{and } \mathrm{H}(P_t)\le\tau_{\mathrm{h}}, \\
0 & \text{otherwise},
\end{cases}
\end{equation}
where $\mathrm{H}(P_t) = -\sum_{v} P_t(v)\log P_t(v)$ is the entropy,
$\mathrm{VE}(P_t)=\sum_{v} P_t(v)\big(-\log P_t(v) - \mathrm{H}(P_t)\big)^2$ is the varentropy,
and $\tau_{\mathrm{v}}$ and $\tau_{\mathrm{h}}$ are the thresholds of varentropy and entropy to control the tradeoff between decoding exploration and computational cost.

The branching function $F(x,\xi_t)$ is then defined as
\begin{equation}
\label{eq:new_branch}
F(x,\xi_t) =
\begin{cases}
\{v_1,\dots,v_K\} & \text{if } \delta_{\text{dec}}(P_t) = 1, \\
\{v_1\},\; v_1\sim P_t & \text{otherwise},
\end{cases}
\end{equation}
where $\{v_1,\dots,v_K\}$ denotes the top-$K$ probable tokens under $P_t$. Rather than branching at every step, \Algnameabbr{} selectively generates new branches only at decision tokens~($\delta_{\text{dec}}(P_t) \!=\! 1$), where the model assigns substantial probability mass to a small set of alternative continuations. In all other cases, when the prediction is either clearly deterministic or broadly uncertain~($\delta_{\text{dec}}(P_t) \!=\! 0$), \Algnameabbr{} refrains from unnecessary branching and conserves space. In the extreme case $\tau_{\mathrm{v}} \to +\infty$, \Algnameabbr{} reduces to standard auto-regressive decoding with a single token generated at each step.

\paragraph{Parallel Branch Expansion.}
\Algnameabbr{} performs auto-regressive generation across all branches in parallel, as illustrated in Figure~\ref{fig:acc-vs-tokens}~(a). 
At each time step $t$, \Algnameabbr{} maintains a batch of reasoning trajectories $\mathcal{T}_t = \{\xi_t^1, \xi_t^2, \dots\}$, initialized with $\mathcal{T}_0 = {\varnothing}$. 
For every trajectory $\xi_t^i \in \mathcal{T}_t$, the model generates the next tokens based on the branch function $F(x, \xi_t^i)$ from Equation~(\ref{eq:new_branch}), and attaches them to form new trajectories $\{\xi_t^i \oplus v_j \mid v_j \in F(x, \xi_t^i)\}$. 
Consequently, the reasoning set is updated as
\begin{equation}
\label{eq:branch-autoregressive}
    \mathcal{T}_{t+1} = \{ \xi^i_t \oplus v_j ~|~  v_j \in F(x, \xi^i_t), \xi^i_t \in \mathcal{T}_t \}.
\end{equation}
This process iterates for $t = 0, 1, 2, \dots$, progressively expanding all branches following Equation~(\ref{eq:branch-autoregressive}). 

\begin{figure*}[t]
\vspace{-2mm}
\centering
\includegraphics[width=0.9\textwidth]{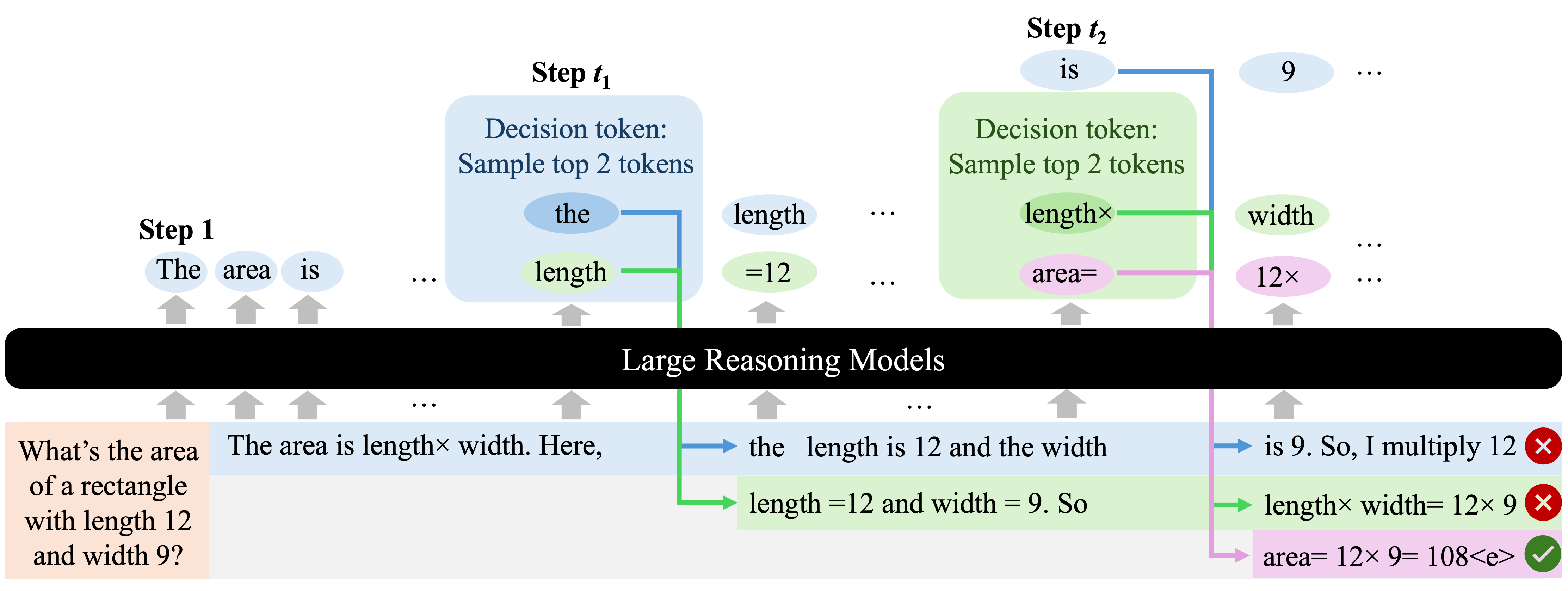}
\caption{\label{fig:dts-decoding-example}An example of \texttt{DTS-Greedy} decoding process with $K=2$, given the input prompt 'What’s the area of a rectangle with length 12 and width 9?'. \Algnameabbr{} selectively generates new branches at steps $t_1$ and $t_2$, and terminates as soon as any branch terminates with an ending token. The final output is \textcolor{blue}{'The area is length× width. Here, length =12 and width = 9. So area= 12× 9= 108'.}} 
\vspace{-1mm}
\end{figure*}

\subsection{Early Termination Strategy}
The early termination strategy in \Algnameabbr{} is motivated by the anti-correlation in Figure~\ref{fig:anti-correlation}, where the short reasoning process consistently outperforms the long reasoning process.
Specifically, \Algnameabbr{} provides two termination strategies: \texttt{DTS-Greedy} and \texttt{DTS-Stable}.
\vspace{-1mm}
\paragraph{DTS-Greedy.} 
\texttt{DTS-Greedy} follows a greedy stopping rule, terminating as soon as any candidate trajectory completes the generation. At each decoding step, \Algnameabbr{} monitors all active reasoning trajectories within the decoding tree to identify if any have generated the end-of-sequence \texttt{EOS} token. If at least one trajectory concludes, the generation process terminates immediately and this completed trajectory is returned as the final reasoning trace and answer.
\vspace{-1mm}
\paragraph{DTS-Stable.} 
\texttt{DTS-Stable} relaxes the greedy behavior in order to obtain more robust decisions. Instead of immediately committing to the first terminating trajectory, \texttt{DTS-Stable} collects a voting budget of $B$ trajectories that have finished and employs majority voting to select a final answer among them. In this way, \texttt{DTS-Stable} preserves the early-completion principle motivated by the length-accuracy anti-correlation, while yielding decisions that are more stable and less sensitive to stochastic variation across individual branches.
\vspace{-2mm}
\paragraph{An Example.}
An illustrative example of \texttt{DTS-Greedy} is shown in Figure~\ref{fig:dts-decoding-example}.
Given the input prompt: \textcolor{blue}{'What’s the area of a rectangle with length 12 and width 9?'}
From step 1 to $t_1-1$, the next-token distribution satisfies $\delta_{\text{dec}}(P_t) =0$, thus \Algnameabbr{} samples a single token per step to produce the prefix \textcolor{blue}{'The area is length× width.'}~(blue).
At step $t_1$, after feeding the token \textcolor{blue}{'Here,'} into the model, the next-token distribution meets $\delta_{\text{dec}}(P_{t_1}) =1$; \Algnameabbr{} therefore generates new branches~(green) by selecting the top two tokens \textcolor{blue}{'the'} and \textcolor{blue}{'length'}, where each branch starts with these two tokens.
From step $t_1+1$ to $t_2-1$, \Algnameabbr{} proceeds each branch~(blue and green) with single-token sampling due to $\delta_{\text{dec}}(P_t) =0$. 
At step $t_2$, after feeding the token \textcolor{blue}{'So'} from the green branch, the condition $\delta_{\text{dec}}(P_{t_2}) =1$ holds again, and \Algnameabbr{} expands the green branch by selecting the top two tokens \textcolor{blue}{'length$\times$'} and \textcolor{blue}{'area='}, yielding three branches in total~(blue, green, and purple).
After step $t_2$, decoding continues with single-token sampling along all active branches and stops as soon as any branch emits the end token \texttt{EOS}~(purple branch).
The final output is \textcolor{blue}{'The area is length× width. Here, length =12 and width = 9. So area= 12× 9= 108'.} 

\vspace{-2mm}
\subsection{The Algorithm of \Algnameabbr{}}
\Algnameabbr{} is described in Algorithm~\ref{alg:dts}.
The algorithm begins by initializing the reasoning set and the completed trajectories set with $\varnothing$~(line~1).
During the decoding, \Algnameabbr{} follows Equation~(\ref{eq:new_branch}) to expands new branches~(line~3); and then follows Equation~(\ref{eq:branch-autoregressive}) to update the reasoning process~(line~4).
At each step, any trajectories that have reached the ending token are added to the completed set~(line~5).
Decoding continues until the size of the completed trajectories set has reached the voting budget, at which point a final voted answer is returned~(lines~6-8). For \texttt{DTS-Greedy}, the voting budget is one.
Overall, \Algnameabbr{} follows a breadth-first search strategy over the sketched decoding tree, ensuring that short and completed reasoning trajectories are prioritized under both \texttt{DTS-Greedy} and \texttt{DTS-Stable}.

\begin{table*}[h!]
    \vspace{-1mm}
    \centering
    \caption{Accuracy (\%) $\uparrow$ of \Algnameabbr{} compared to baselines on the AIME24, AIME25, GPQA-D, and LiveBench datasets.}
    \resizebox{\textwidth}{!}{
    \scriptsize
    \begin{tabular}{l|l|cc|cc|cc|cc|cc}
    \toprule
         & & \multicolumn{2}{c|}{AIME24} & \multicolumn{2}{c|}{AIME25} & \multicolumn{2}{c|}{GPQA-D} & \multicolumn{2}{c|}{LiveBench} & \multicolumn{2}{c}{Average} \\
    \midrule
         Model & Method & ACC & Improve & ACC & Improve & ACC & Improve & ACC & Improve & ACC & Improve \\
    \midrule
        \multirow{6}{*}{DS-Distill-Qwen-1.5B} 
        & Standard inference         & 26.67 & 0.00 & 24.67 & 0.00 & 32.02 & 0.00 & 6.00 & 0.00 & 22.34 & 0.00\\
        & Self-Consistency      & 40.67 & +14.00 & 33.33 & +8.66 & 37.58 & +5.56 & 8.00 & +2.00 & 29.89 & +7.55 \\
        & DeepConf-low    & 46.00 & +19.33 & 34.00 & +9.33 & 38.89 & +6.87 & 9.30 & +3.30 & 32.05 & +9.71 \\
        & DeepConf-high    & 44.00 & +17.33 & \underline{36.00} & \underline{+11.33} & 36.87 & +4.85 & 12.00 & +6.00 & 32.22 & +9.88 \\
        & \texttt{DTS-Greedy}    & \underline{54.67} & \underline{+28.00} & 34.67 & +10.00 & \textbf{41.41} & \textbf{+9.39} & \textbf{17.30} & \textbf{+11.30} & \underline{37.01} & \underline{+14.67} \\
        & \texttt{DTS-Stable}    & \textbf{64.67} & \textbf{+38.00} & \textbf{39.33} & \textbf{+14.66} & \underline{41.11} & \underline{+9.09} & \underline{16.70} & \underline{+10.70} & \textbf{40.45} & \textbf{+18.11}\\
        \midrule
       \multirow{6}{*}{DS-Distill-Qwen-7B} 
        & Standard inference         & 52.67 & 0.00 & 36.00 & 0.00 & 49.29 & 0.00 & 27.20 & 0.00 & 41.29 & 0.00\\
        & Self-Consistency      & 69.33 & +16.66 & \underline{54.00} & \underline{+18.00} & 53.03 & +3.74 & \underline{36.70} & \underline{+9.50} & 53.27 & +11.97 \\
        & DeepConf-low    & 70.67 & +18.00 & 52.67 & +16.67 & 52.32 & +3.03 & 33.30 & +6.10 & 52.24 & +10.95 \\
        & DeepConf-high    & 63.33 & +10.66 & 49.33 & +13.33 & 52.32 & +3.03 & 28.70 & +1.50 & 48.42 & +7.13 \\
        & \texttt{DTS-Greedy}    & \underline{73.33} & \underline{+20.66} & 53.33 & +17.33 & \underline{55.76} & \underline{+6.47} & 32.70 & +5.50 & \underline{53.78} & \underline{+12.49} \\
        & \texttt{DTS-Stable}    & \textbf{78.67} & \textbf{+26.00} & \textbf{56.67} & \textbf{+20.67} & \textbf{57.78} & \textbf{+8.49} & \textbf{40.70} & \textbf{+13.50} & \textbf{58.45} & \textbf{+17.16}\\
        \midrule
        \multirow{6}{*}{Qwen3-0.6B} 
        & Standard inference         & 11.33 & 0.00 & 14.00 & 0.00 & 24.75 & 0.00 & 24.70 & 0.00 & 18.70 & 0.00\\
        & Self-Consistency      & 11.33 & +0.00 & \underline{25.33} & \underline{+11.33} & 25.05 & +0.30 & 27.30 & +2.60 & 22.25 & +3.56 \\
        & DeepConf-low    & 13.33 & +2.00 & 24.00 & +10.00 & \underline{25.96} & \underline{+1.21} & \underline{32.00} & \underline{+7.30} & \underline{23.82} & \underline{+5.13} \\
        & DeepConf-high    & \underline{14.67} & \underline{+3.34} & 20.67 & +6.67 & 25.56 & +0.81 & 22.70 & -2.00 & 20.90 & +2.21 \\
        & \texttt{DTS-Greedy}    & 14.00 & +2.67 & \underline{25.33} & \underline{+11.33} & 25.56 & +0.81 & 28.60 & +3.90 & 23.37 & +4.68 \\
        & \texttt{DTS-Stable}    & \textbf{18.67} & \textbf{+7.34} & \textbf{29.33} & \textbf{+15.33} & \textbf{26.57} & \textbf{+1.82} & \textbf{34.70} & \textbf{+10.00} & \textbf{27.32} & \textbf{+8.62}\\
        \midrule
       \multirow{6}{*}{Phi4-mini-reasoning-4B} 
        & Standard inference         & 49.33 & 0.00 & 36.00 & 0.00 & 50.20 & 0.00 & 42.40 & 0.00 & 44.48 & 0.00\\
        & Self-Consistency      & 61.33 & +12.00 & 42.00 & +6.00 & 52.32 & +2.12 & \underline{55.30} & \underline{+12.90} & 52.74 & +8.26 \\
        & DeepConf-low    & 62.67 & +13.34 & 38.00 & +2.00 & 52.83 & +2.63 & 51.30 & +8.90 & 51.20 & +6.72 \\
        & DeepConf-high    & 50.00 & +0.67 & 35.33 & -0.67 & 51.72 & +1.52 & 38.70 & -3.70 & 43.94 & -0.55 \\
        & \texttt{DTS-Greedy}    & \underline{68.00} & \underline{+18.67} & \underline{44.00} & \underline{+8.00} & \textbf{53.13} & \textbf{+2.93} & 52.80 & +10.40 & \underline{54.48} & \underline{+10.00} \\
        & \texttt{DTS-Stable}    & \textbf{71.33} & \textbf{+22.00} & \textbf{52.00} & \textbf{+16.00} & \underline{53.03} & \underline{+2.83} & \textbf{56.00} & \textbf{+13.60} & \textbf{58.09} & \textbf{+13.61}\\
    \bottomrule
    \end{tabular}
    }
    \label{tab:exp_acc}
    \vspace{-3mm}
\end{table*}

\begin{algorithm}[H]
\caption{\Algname{} (\Algnameabbr{})}
\label{alg:dts}

{\bfseries Input:} LRM $f$, input prompt $x$, voting budget $B$.\\
{\bfseries Output:} Final reasoning answer $y^*$.

\begin{algorithmic}[1]

\STATE $\mathcal{T}_0, \mathcal{C} = \varnothing, \varnothing$

\FOR{$t = 1, 2, \cdots$}

\STATE Generate new branches by Eq.~(\ref{eq:new_branch})

\STATE Collect trajectory candidates $\mathcal{T}_{t}$ by Eq.~(\ref{eq:branch-autoregressive})

\STATE $\mathcal{C} = \mathcal{C} \cup\{\xi^i_t \in \mathcal{T}_t ~|~ \texttt{EOS} \in \xi^i_t\}$

\IF{$|\mathcal{C}|$ $\ge$ $B$}
    \STATE $y^*\gets$ Majority voting of candidate answers in $\mathcal{C}$
    \STATE \textbf{return} $y^*$
\ENDIF

\ENDFOR

\end{algorithmic}
\end{algorithm}
\section{Experiments}
In this section, we conduct experiments to evaluate the performance of \Algnameabbr{} framework, aiming to answer the following research questions: \textbf{RQ1}: Does \Algnameabbr{} produce more accurate reasoning? \textbf{RQ2}: Can \Algnameabbr{} mitigate the repetitive generation during reasoning? \textbf{RQ3:} Does \Algnameabbr{} select more reasonable trajectories?
\subsection{Experimental Setup}
We specify the models, datasets, and baseline methods below. We provide more information on the experiment implementation details in Appendix~\ref{sec:imp_det}.
\paragraph{Models.} We evaluate \Algnameabbr{} with four representative LRMs:  DeepSeek-R1-Distill-Qwen-7B and -1.5B~\cite{guo2025deepseek}, Qwen3-0.6B~\cite{yang2025qwen3}, and Phi4-mini-reasoning-4B~\cite{abdin2024phi}, whose pre-trained weights are from Huggingface Transformers~\cite{wolf2020transformers}.

\paragraph{Datasets.} The evaulation of \Algnameabbr{} is based on four reasoning dataset: AIME24~\cite{aime24}, AIME25~\cite{aime25}, GPQA-Diamond~\cite{rein2024gpqa}, and LiveBench-Reasoning~\cite{white2024livebench}. All datasets are accessed through the Huggingface Datasets~\cite{lhoest2021datasets} library. AIME24 and AIME25 consist of high-difficulty problems from the American Invitational Mathematics Examination, each containing 30 questions. 
GPQA-Diamond contains 198 graduate-level STEM questions.
LiveBench-Reasoning contains 200 questions on logical deduction, puzzle solving, and spatial reasoning.
We follow existing works~\cite{chen2025verithinker, xu2025self} to construct prompts and extract answers for all datasets.

\vspace{-4mm}
\paragraph{Baseline Methods.}
\textbf{Standard Inference:}  We evaluate each LRM under its default inference setting. The model generates a single reasoning trace for each problem, and the final predicted answer is compared against the ground-truth label from the dataset.
\textbf{Self-Consistency:} Self-Consistency~\cite{wang2022self} allows the LRM to generate multiple reasoning trajectories independently for each problem. Each trajectory produces a candidate answer, and the final prediction is obtained by majority voting over these answers.
\textbf{DeepConf:} DeepConf~\cite{fu2025deep} improves multi-trajectory reasoning by using confidence
scores from the model’s next-token distributions to filter out low-quality traces during generation. The
method first estimates a confidence threshold and continues generation while discarding traces whose confidence falls below this threshold. We evaluate DeepConf-low, which applies a stricter threshold and keeps only highly confident traces, and  DeepConf-high, which uses a looser threshold to encourage exploration.

\vspace{-2mm}
\subsection{Performance on Reasoning Tasks~(RQ1)}

Table~\ref{tab:exp_acc} shows the accuracy(\%) of \Algnameabbr{}, compared with baseline methods and standard inference of LRMs.
\vspace{-3mm}
\paragraph{Accuracy Improvement.} As shown in Table~\ref{tab:exp_acc}, \Algnameabbr{} consistently outperforms all baseline methods and standard inference. Furthermore, the single-trajectory variant \texttt{DTS-Greedy} surpasses multi-trajectory baselines such as Self-Consistency and DeepConf in most cases, indicating that selectively branching at decision tokens is more effective than generating more independent samples. These results show that by sketching the decoding tree and with early termination, \Algnameabbr{} steers decoding toward existing optimal solutions in the reasoning space.
\vspace{-2mm}
\paragraph{Model Agnosticism.} By integrating with different families of LRMs, \Algnameabbr{} delivers consistent performance, as shown in Table~\ref{tab:exp_acc}. Since it operates purely at inference time and requires no model post-training, its improvements transfer across different model architectures. This indicates that \Algnameabbr{} can potentially serve as a versatile enhancement for a range of LRMs in practice.
\vspace{-2mm}
\paragraph{Task Generalization.} 
Beyond math, \Algnameabbr{} demonstrates strong performance across diverse reasoning domains. \Algnameabbr{} exhibits strong and stable accuracy gains in both GPQA-D and LiveBench in Table~\ref{tab:exp_acc}. The domains from these datasets span across physics, chemistry, biology, semantic understanding, and logical reasoning. These robust gains indicate \Algnameabbr{} provides a general enhancement across diverse subjects and tasks. 
\begin{table}[!t]
    \centering
    \caption{Repetition rate (\%) $\downarrow$ of Standard Inference and \texttt{DTS-Greedy} on different models and datasets.}
    \label{tab:repetition-rate}
    \scriptsize
    \resizebox{\columnwidth}{!}{%
    \begin{tabular}{l l cc}
        \toprule
        \textbf{Model} & \textbf{Dataset} & \textbf{Std. Inf.} & \textbf{DTS-Greedy} \\
        \midrule
        \multirow{5}{*}{DS-Distill-Qwen-1.5B}
            & AIME24     & 15.33 & \textbf{4.67} \\
            & AIME25     & 26.67 & \textbf{6.00} \\
            & GPQA-D     & 1.40  & \textbf{0.00} \\
            & LiveBench  & 26.40 & \textbf{0.00} \\
            & Average    & 17.45 & \textbf{2.67} \\
        \midrule
        \multirow{5}{*}{DS-Distill-Qwen-7B}
            & AIME24     & 6.67  & \textbf{1.33} \\
            & AIME25     & 12.67 & \textbf{2.67} \\
            & GPQA-D     & 0.50  & \textbf{0.00} \\
            & LiveBench  & 17.20 & \textbf{0.00} \\
            & Average    & 9.26  & \textbf{1.00} \\
        \midrule
        \multirow{5}{*}{Qwen3-0.6B}
            & AIME24     & 5.33  & \textbf{0.00} \\
            & AIME25     & 4.00  & \textbf{0.00} \\
            & GPQA-D     & 1.00  & \textbf{0.00} \\
            & LiveBench  & 2.00  & \textbf{0.00} \\
            & Average    & 3.08  & \textbf{0.00} \\
        \midrule
        \multirow{5}{*}{Phi4-mini-reasoning-4B}
            & AIME24     & 6.00  & \textbf{0.00} \\
            & AIME25     & 9.33  & \textbf{0.00} \\
            & GPQA-D     & 1.20  & \textbf{0.00} \\
            & LiveBench  & 2.80  & \textbf{0.00} \\
            & Average    & 4.83  & \textbf{0.00} \\
        \bottomrule
    \end{tabular}%
    \vspace{-2mm}
    }
\end{table}

\begin{table}[t]
    \centering
    \caption{Accuracy~(\%) $\uparrow$ comparison of confidence-based log-probability trajectory selection and \Algnameabbr{}'s early stopping selection on different models and datasets.}
    \label{tab:selection}
    \scriptsize
    \resizebox{\columnwidth}{!}{%
    \begin{tabular}{l l cc}
        \toprule
        \textbf{Dataset} & \textbf{Model} & \textbf{Max Logits} & \textbf{DTS-Greedy} \\
        \midrule
        \multirow{4}{*}{AIME24}
            & DS-Distill-Qwen-1.5B       & 38.67 & \textbf{54.67} \\
            & DS-Distill-Qwen-7B         & 65.33 & \textbf{73.33} \\
            & Qwen3-0.6B                 & 7.33 & \textbf{14.00} \\
            & Phi4-mini-reasoning-4B     & 58.00 & \textbf{68.00} \\
        \midrule
        \multirow{4}{*}{AIME25}
            & DS-Distill-Qwen-1.5B       & 21.33 & \textbf{34.67} \\
            & DS-Distill-Qwen-7B         & 50.00 & \textbf{53.33} \\
            & Qwen3-0.6B                 & 16.67 & \textbf{25.33} \\
            & Phi4-mini-reasoning-4B     & 42.00 & \textbf{44.00} \\
        \bottomrule
    \end{tabular}%
    }
    \vspace{-4mm}
\end{table}

\subsection{Mitigation of Repetitive Reasoning~(RQ2)}
\vspace{-1mm}
\label{sec:repetition}
In this section, we show that \Algnameabbr{} reduces repetition in LRM decoding. The repetition problem is a special case of overthinking where the model falls into a reasoning loop and continuously generates repeating phrases or tokens without concluding. This phenomenon signals degraded reasoning quality, as the model stagnates on intermediate steps and accumulates inconsistencies~\cite{yao2025understanding, pipis2025wait}. Additionally, as recent studies~\cite{xie2025word} indicate, once repetition occurs, LRMs are unlikely to untrap themselves and exploit all decoding token budget with no final answer. \Algnameabbr{} provides a solution to this failure mode intrinsically by sketching the reasoning tree and favoring the shorter completed trajectories. Trajectories that fall into repetition are overridden by concise completions, preventing the decoder from returning repeated segments.

Table~\ref{tab:repetition-rate} reports the rate of cases where repetition occurred under standard inference and \texttt{DTS-Greedy}, where the repetition is identified by the LRM reaching the maximum token budget. Across all datasets and models, \texttt{DTS-Greedy} consistently reduces the repetition rate, demonstrating its ability to recover from endless generation. These results confirm \Algnameabbr{} sketches toward more reliable and consistent optimal solutions. 
\begin{figure}[t]
    \centering
    \begin{subfigure}[t]{1\columnwidth}
        \centering
        \includegraphics[width=\linewidth]{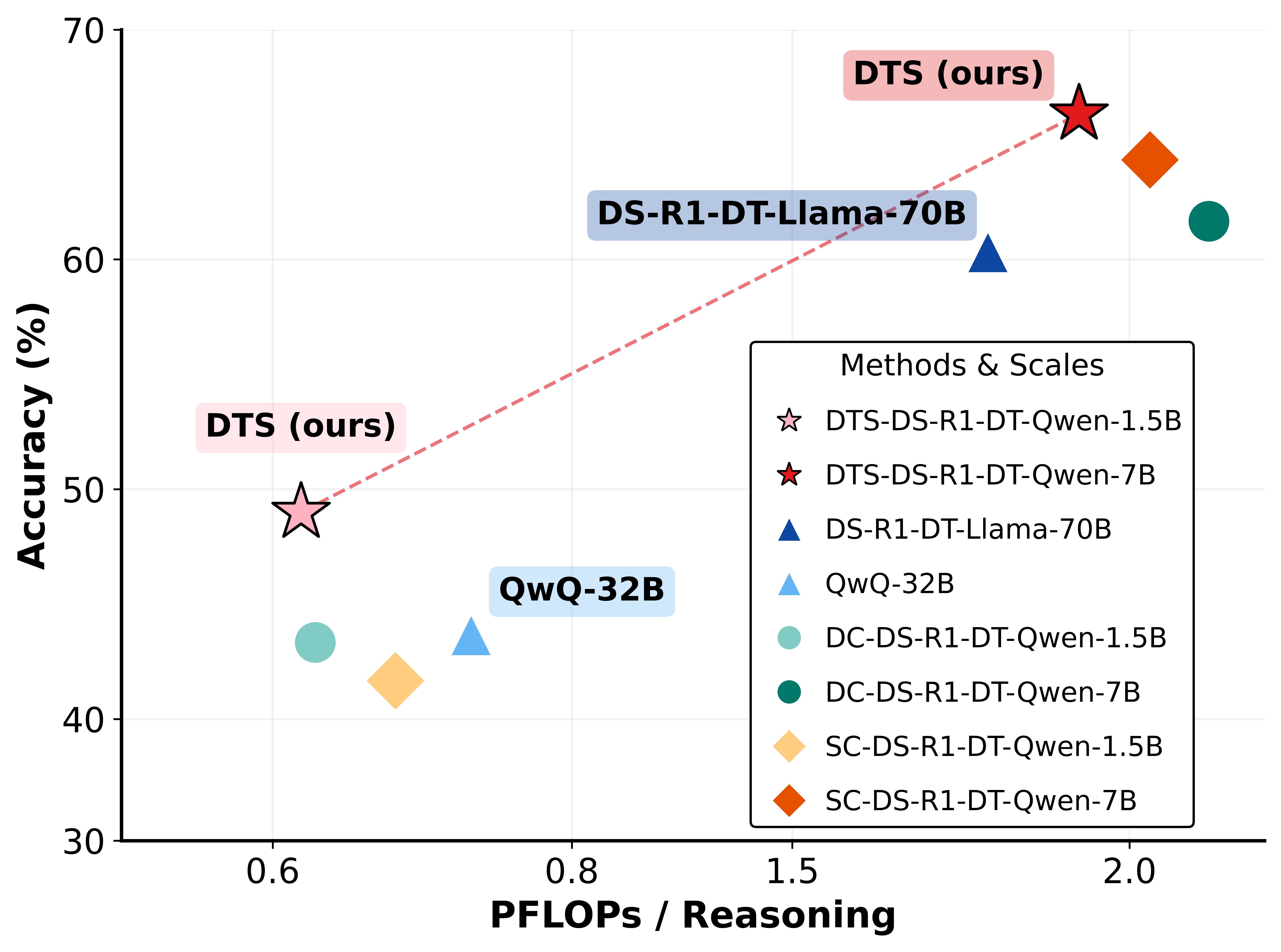}
    \end{subfigure}
    \caption{\label{fig:dts-scaling}
    Average Accuracy~(\%) $\uparrow$ vs.\ PFLOPs $\downarrow$ of \Algnameabbr{} and baseline methods compared with flagship LRMs DS-Distill-Llama-70B and QwQ-32B-Preview on the AIME24 and AIME25 datasets.}
    \vspace{-5mm}
\end{figure}

\begin{figure*}[t]
\vspace{-2mm}
\centering
\includegraphics[width=0.9\textwidth]{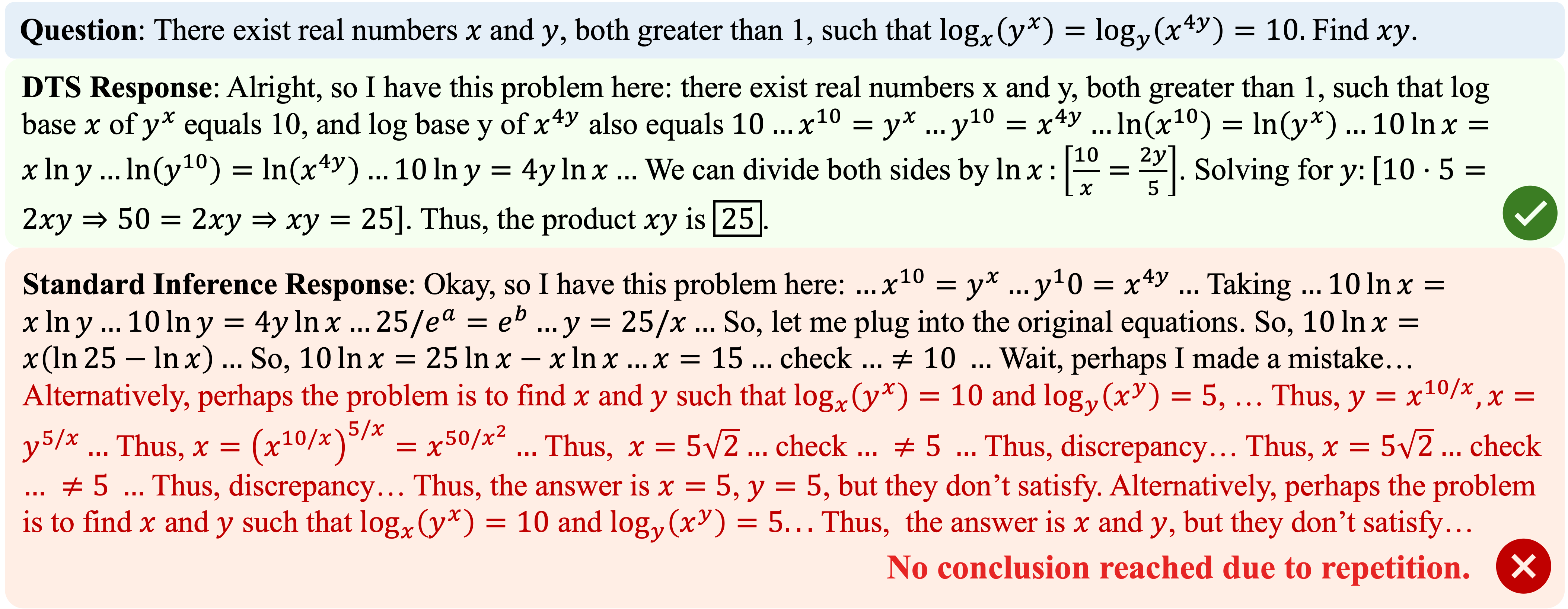}
\caption{\label{fig:case_study}A case study illustrating how \Algnameabbr{} produces a correct solution, while standard inference enters an endless repetition loop, failing to reach a conclusion after consuming all token budgets.}
\vspace{-3mm}
\end{figure*}
\vspace{-1mm}
\subsection{DTS versus Max-logit Trajectory Selection}
\vspace{-1mm}
We compare our shortest trajectory selection with confidence-based selection for the final answer in this section. Specifically, a common practice for confidence-based selection is to use LRM's cumulative logit as the decision indicator~\cite{leang2025picsar, just2025distilling, xu2025self}, selecting the trajectory that the model assigns the highest confidence. However, this assumption has been widely challenged in recent studies~\cite{simhi2025trust, kumar2024confidence, steyvers2025large}, showing that models can assign high probability to fluent yet logically incorrect traces, and that confidence can increase with repetition or local consistency rather than global correctness. In contrast, our shortest trajectory selection strategy is motivated by both the theoretical aspects of the LRM's post-training scheme and the empirically observed anti-correlation between reasoning length and accuracy in Section~\ref{sec:overthinking}. Table~\ref{tab:selection} presents the performance comparison between confidence-based max-logit selection and shortest-trajectory selection across all models and datasets\footnote{We follow the Beam Search method~\cite{wu2016google} to support max-logit selection with length normalization.}. We observe that our selection strategy consistently achieves higher accuracy in every setting. These results support that selecting the shortest completed trace is a better criterion than confidence-based scoring.

\subsection{Improving Inference-time Scaling}
\Algnameabbr{} achieves better inference-time scaling by delivering higher performance per unit of compute. 
We report the average accuracy and computational footprint for all models and baselines on AIME24 and AIME25 in Figure~\ref{fig:dts-scaling}. 
\vspace{-2mm}
\paragraph{Comparison with Flagship LRMs.} We first compare smaller LRMs utilizing \texttt{DTS-Stable} with industry-scale flagship LRMs under matched compute budgets. At similar PFLOPs levels, DS-Distill-Qwen-1.5B and 7B achieve higher accuracy than QwQ-32B-Preview and DS-Distill-Llama-70B, respectively. This demonstrates that \Algnameabbr{} enhances smaller LRMs to reach flagship-level reasoning performance under identical inference-scaling conditions.
\vspace{-2mm}
\paragraph{Comparison with Parallel Thinking Baselines.}
We further compare \Algnameabbr{} with existing parallel thinking methods applied to the same LRMs. \Algnameabbr{} consistently achieves higher accuracy at lower PFLOPs than Self-Consistency and DeepConf. These results highlight the effectiveness of \Algnameabbr{}'s structured tree sketching and reliable trajectory selection, yielding a more favorable accuracy-compute scaling than baseline methods. 

\subsection{Demonstration of \Algnameabbr{} Reasoning~(RQ3)}
To further illustrate how \Algnameabbr{} enhances reasoning behavior, we present a case study in
Figure~\ref{fig:case_study} from the AIME25 dataset using the DS-Distill-Qwen-1.5B model. Under standard inference, the
model repeatedly revisits the same partial calculations ultimately failing to reach a conclusion. This behavior corresponds to the repetitive
reasoning phenomenon discussed in Section~\ref{sec:repetition}, where the repeating segments are
highlighted in \textcolor{red}{red}. 
In contrast, \Algnameabbr{} sketches the decoding tree around decision tokens and settles on a coherent reasoning trajectory that leads to the correct answer. The final trace remains concise, logically structured, and aligned with the correct solution, illustrating how \Algnameabbr{} helps the model avoid reasoning loops and converge on a reliable trajectory. We provide more case studies in Appendix~\ref{sec:more_case}.

\section{Related Works} 
\paragraph{Parallel Thinking.} 
Parallel thinking methods, such as Self-Consistency~\cite{wang2022self}, Best-of-N~\cite{brown2024large}, and DeepConf~\cite{fu2025deep}, enhance LRMs by generating multiple reasoning trajectories and aggregating their solutions through a voting or scoring mechanism. By exploring various reasoning, these methods reduce the impact of individual reasoning errors and improve overall inference robustness. As a result, parallel thinking has become a widely adopted inference-time technique for improving LRM reasoning performance. 
\vspace{-2mm}
\paragraph{Adaptive Thinking.}
Another line of research studies supervised fine-tuning (SFT) and RL methods that adapt reasoning length during the training phase. For instance, AutoL2S~\cite{luo2025autol2s}, DAST~\cite{shen2025dast}, and L1~\cite{aggarwal2025l1} optimize models to generate either short or long responses based on the perceived difficulty of the prompt. However, these adaptive thinking methods incur additional training and data curation costs and are unable to improve LRM reasoning accuracy.

\vspace{-3mm}
\section{Conclusion}
\vspace{-1mm}

In this work, we introduced \Algname{} (\Algnameabbr{}), a training-free decoding framework for enhancing LRM reasoning. \Algnameabbr{} enables structured reasoning explorations by sketching a compact decoding tree and leveraging the length-accuracy anti-correlation to select short yet reliable reasoning for the final solution.
\Algnameabbr{} branches selectively at decision tokens to maintain trajectory diversity. 
Motivated by a proven length-accuracy anti-correlation, \Algnameabbr{} prioritizes high-quality solutions via early termination.
Experiments across four models and four datasets comprehensively show stable accuracy gains and mitigation of repetition. 
These results show that \Algnameabbr{} enhances LRM reasoning through structured exploration and reasoning selection.

\section*{Impact Statement}
This paper presents work whose goal is to advance the field of Machine
Learning. There are many potential societal consequences of our work, none
which we feel must be specifically highlighted here.

\bibliography{references}
\bibliographystyle{icml2026}

\newpage
\appendix
\onecolumn
\section{Implementation Details}
\label{sec:imp_det}
For \Algnameabbr{}, we set varentropy threshold $\tau_{\mathrm{v}} = 1.5$, entropy threshold $\tau_{\mathrm{h}} = 2.5$, and $K=3$ across all datasets and models. To prevent the decoding tree from growing exponentially, we additionally limit the number of simultaneously active branches at 48. Once this limit is met, \Algnameabbr{} temporarily disables branching and always applies the second case of Equation~\ref{eq:new_branch}, even if $\delta_{\text{dec}}(P_t)=1$. This design prioritizes earlier decision tokens that determine the high-level shape of the sketched decoding tree, while suppressing later ones that mostly dive into minor variations of already-established branches. For multi-trajectory methods with majority voting, including Self-Consistency, DeepConf, and \texttt{DTS-Stable}, we set the voting budget $N=8$. For all experiments, we use the LRM's recommended hyperparameter settings and maximum token budget. Stochasticity is controlled by fixing random seeds $s \in \{0,1,2,3,4\}$ and the average accuracy over these five runs is reported.
\vspace{-1mm}
\section{More Case Studies}
\begin{figure}[H]
\centering
\includegraphics[width=\textwidth]{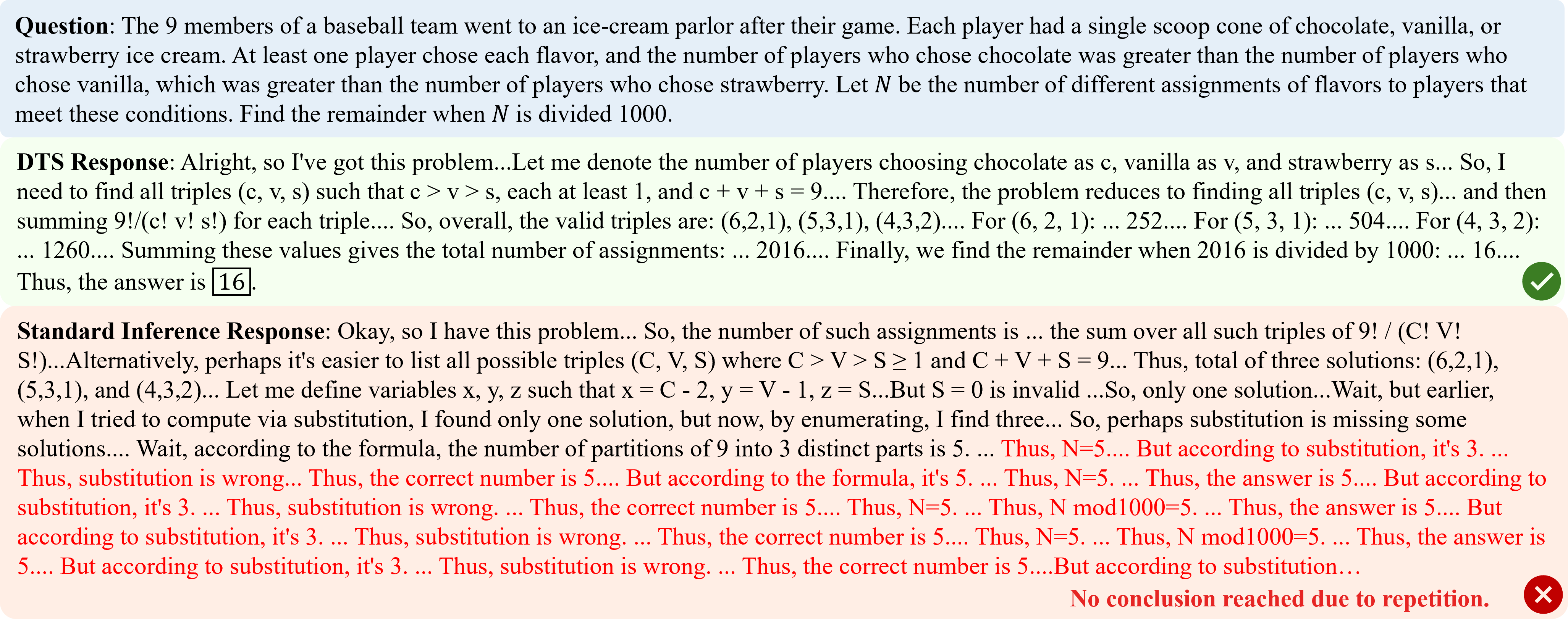}
\caption{\label{fig:case_study_app_1}A case study from AIME25 dataset using DS-Distill-Qwen-1.5B with \Algnameabbr{} and standard inference.}
\end{figure}

\begin{figure}[H]
\centering
\includegraphics[width=\textwidth]{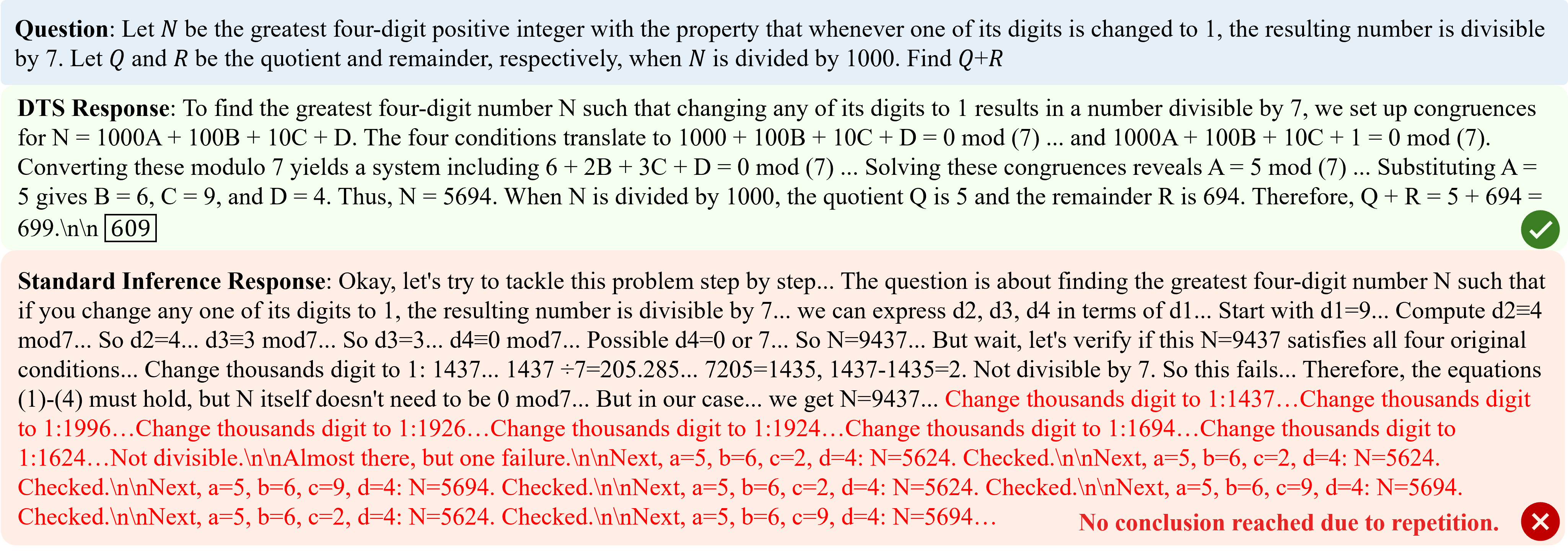}
\caption{\label{fig:case_study_app_5}A case study from AIME24 dataset using Phi4-mini-reasoning-4B with \Algnameabbr{} and standard inference.}
\end{figure}

\begin{figure}[H]
\centering
\includegraphics[width=\textwidth]{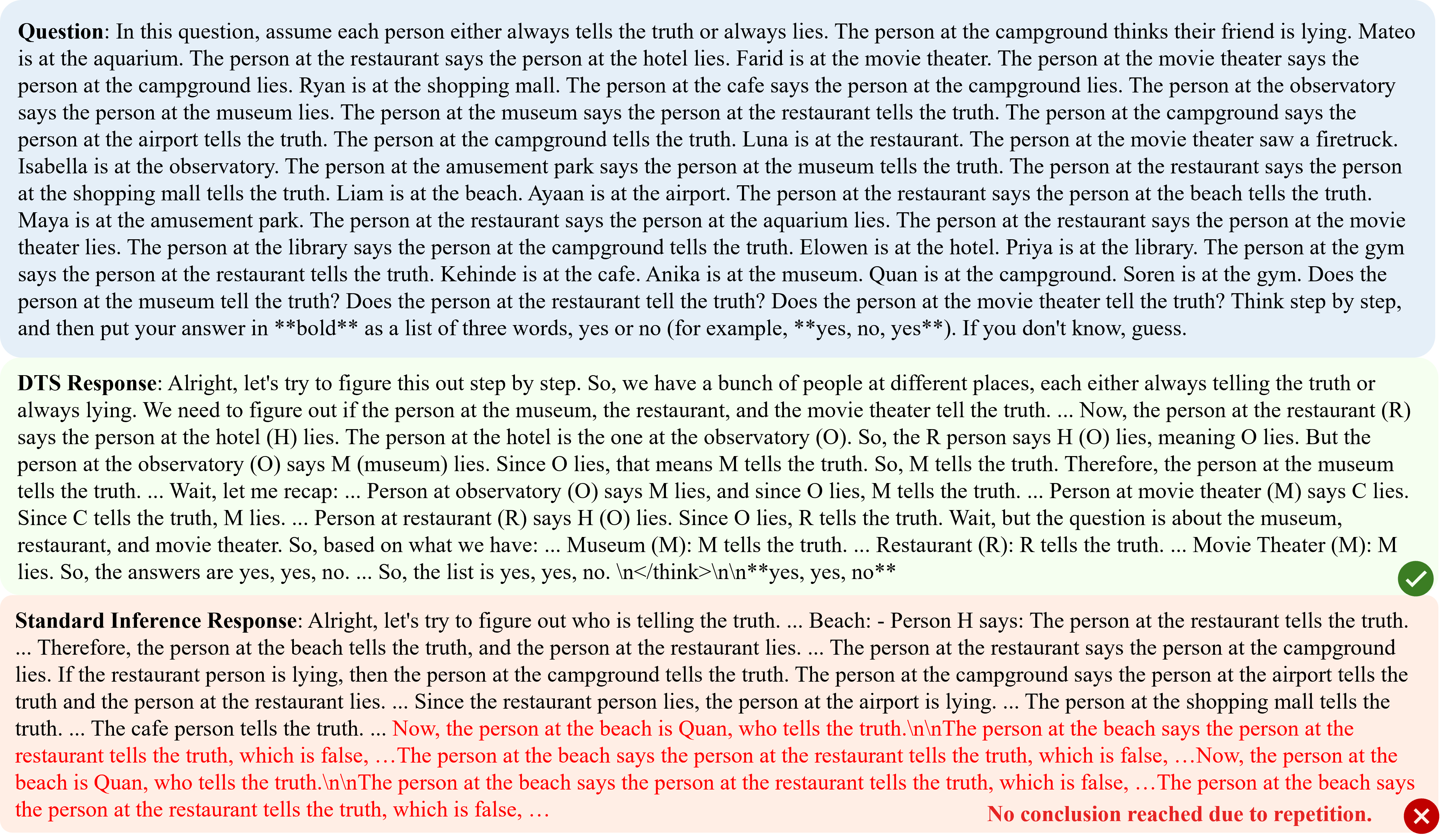}
\caption{\label{fig:case_study_app_2}A case study from LiveBench dataset using DS-Distill-Qwen-1.5B with \Algnameabbr{} and standard inference.}
\end{figure}

\begin{figure}[H]
\centering
\includegraphics[width=\textwidth]{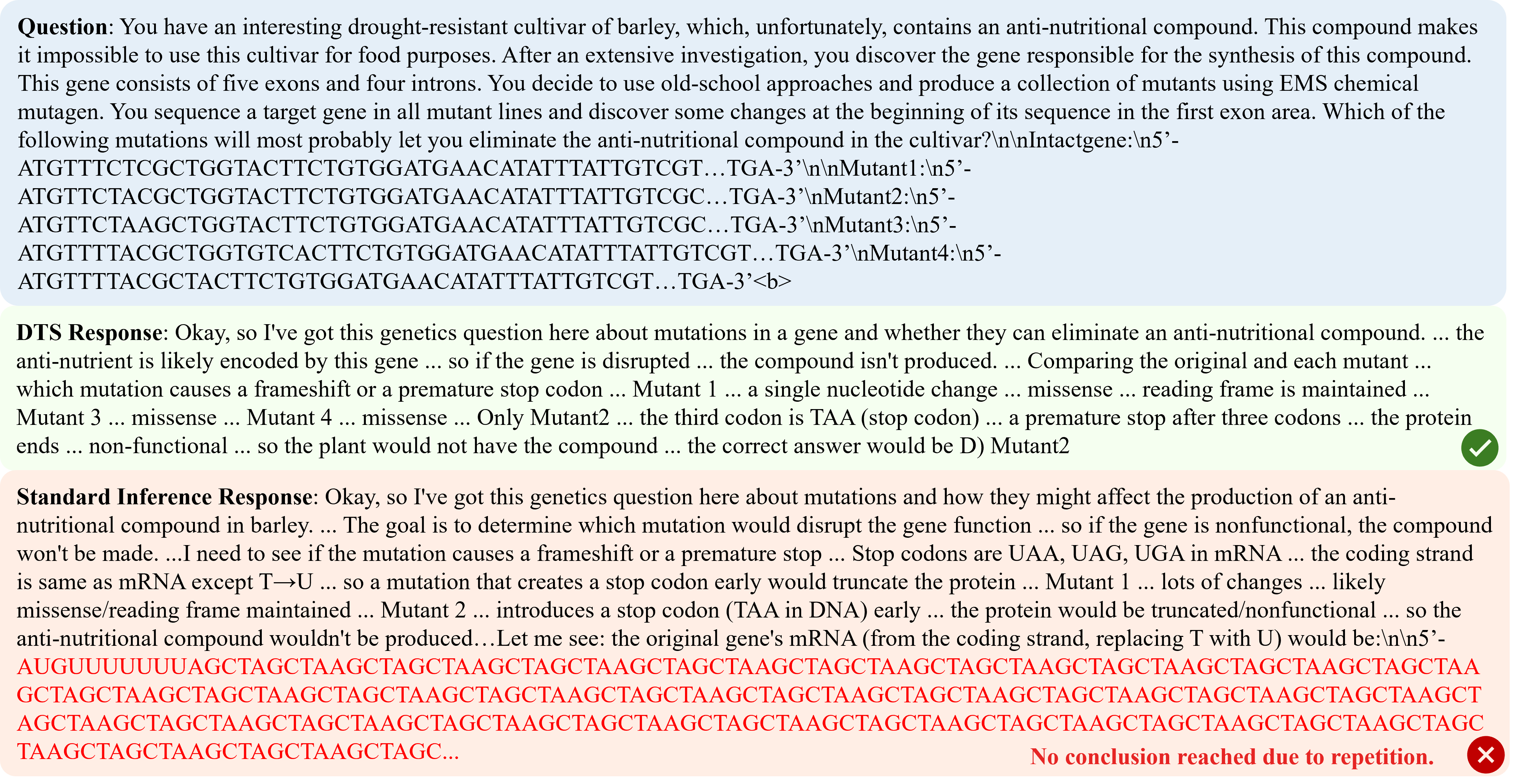}
\caption{\label{fig:case_study_app_3}A case study from GPQA-D dataset using DS-Distill-Qwen-7B with \Algnameabbr{} and standard inference.}
\end{figure}

\begin{figure}[H]
\centering
\includegraphics[width=\textwidth]{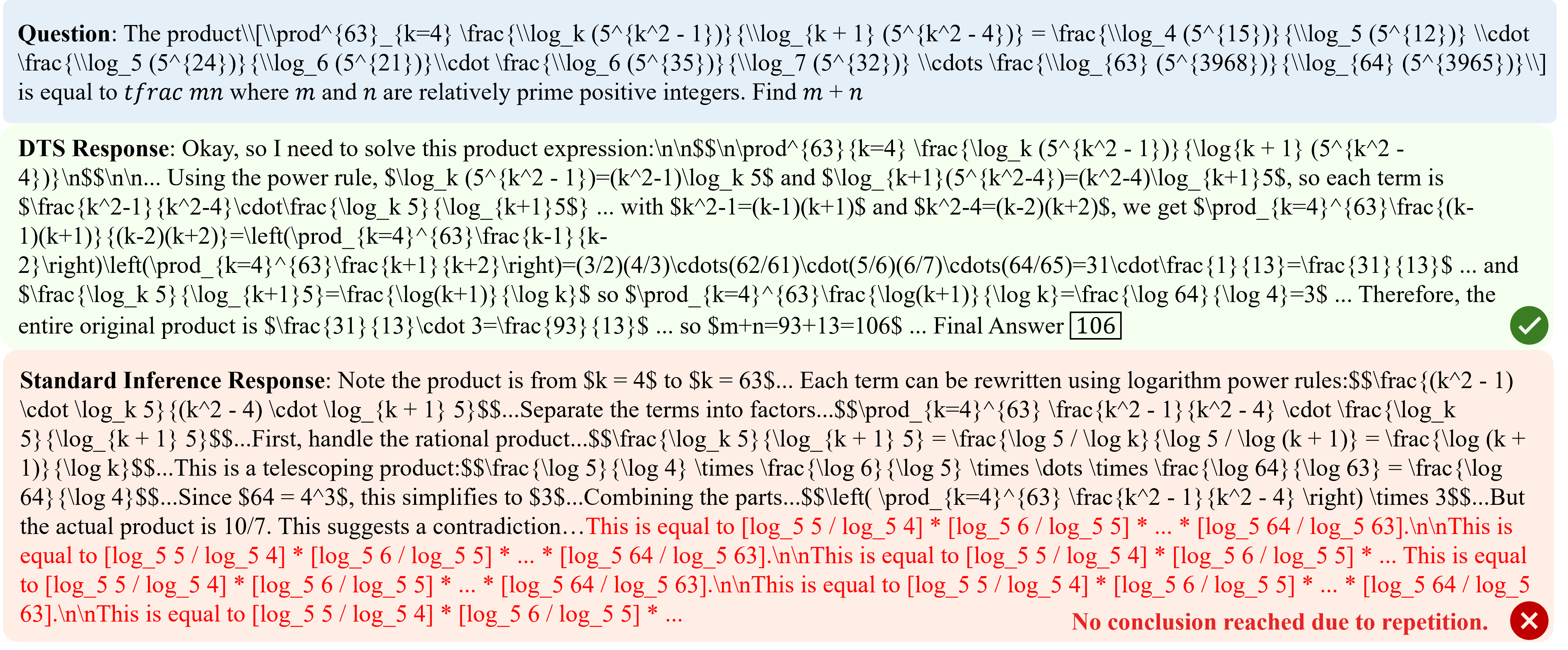}
\caption{\label{fig:case_study_app_4}A case study from AIME25 dataset using Qwen3-0.6B with \Algnameabbr{} and standard inference.}
\end{figure}

\vspace{-1mm}
\label{sec:more_case}
Figure~\ref{fig:case_study_app_1}, ~\ref{fig:case_study_app_5}, ~\ref{fig:case_study_app_2}, ~\ref{fig:case_study_app_3}, and~\ref{fig:case_study_app_4} present additional qualitative case studies spanning all evaluated models and datasets, including AIME24, AIME25, GPQA-D, and LiveBench. Across these diverse reasoning tasks, a consistent pattern emerges: under standard inference, models frequently fall into repetitive or self-contradictory reasoning loops, 
revisiting partial computations or local deductions without converging to a final answer. In contrast, \Algnameabbr{} produces reasoning trajectories that are shorter, better structured, and more aligned with correct solutions. These examples further reinforce the robustness of \Algnameabbr{}, complementing the quantitative improvements reported in Table~\ref{tab:exp_acc}.
\vspace{-2mm}
\vspace{-3mm}

\section{Decision Token Criteria Analysis}

Table~\ref{tab:aime25_decision_token_ablation} compares the performance of using different decision token criteria and threshold values in Equation~\ref{eq:new_branch} on AIME25 with DS-Distill-Qwen-1.5B. We show that \Algnameabbr{}'s low varentropy and high entropy criteria outperform other entropy and varentropy combinations. Furthermore, by altering the entropy threshold, the performance of \Algnameabbr{} stays relatively stable.  

\begin{table}[H]
\centering
\small
\setlength{\tabcolsep}{8pt}
\renewcommand{\arraystretch}{1.15}
\caption{Accuracy~(\%) $\uparrow$ comparison of different decision token criteria and threshold values on AIME25 and DS-Distill-Qwen-1.5B.}
\label{tab:aime25_decision_token_ablation}
\begin{tabular}{lccc}
\hline
\textbf{Criteria} &\textbf{Threshold} & \textbf{DTS-Greedy} & \textbf{DTS-Stable} \\
\hline
$\mathrm{H}(P_t)\ge\tau_{\mathrm{h}}$ & $\tau_{\mathrm{h}}{=}2.5$ & 29.33 & 33.33 \\
$(\mathrm{VE}(P_t)\ge\tau_{\mathrm{v}} \ \text{and } \mathrm{H}(P_t)\le\tau_{\mathrm{h}}) \ \text{or}\ \mathrm{H}(P_t)\ge\tau_{\mathrm{h}}$ & $\tau_{\mathrm{v}}{=}1.5,\ \tau_{\mathrm{h}}{=}2.5$ & 30.67 & 34.67 \\
$\mathrm{VE}(P_t)\ge\tau_{\mathrm{v}} \ \text{and } \mathrm{H}(P_t)\le\tau_{\mathrm{h}}$ & $\tau_{\mathrm{v}}{=}1.5,\ \tau_{\mathrm{h}}{=}2.5$ & \textbf{34.67} & \textbf{39.33} \\
$\mathrm{VE}(P_t)\ge\tau_{\mathrm{v}} \ \text{and } \mathrm{H}(P_t)\le\tau_{\mathrm{h}}$ & $\tau_{\mathrm{v}}{=}1.5,\ \tau_{\mathrm{h}}{=}1.2$ & \textbf{34.67} & \underline{38.67} \\
\hline
\end{tabular}
\end{table}

\section{Computational Infrastructure}
The computational infrastructure information is given in Table~\ref{tab:computing_infrastructure}.
\begin{table}[H]
\caption{Experiment configuration and computing infrastructure.}
\centering
\begin{tabular}{l|c}
\toprule
Name & Value \\
\midrule
Data type & \texttt{torch.bfloat16} \\
Flash-Attention & False \\
Eval batch-size & 1 \\
Computing Infrastructure & GPU \\
GPU Model & NVIDIA-H200 \\
GPU Memory & 141GB \\ 
GPU Number & 4 \\
CUDA Version & 12.4 \\
CPU Memory & 512GB \\
\bottomrule
\end{tabular}
\label{tab:computing_infrastructure}
\end{table}


\end{document}